# A Systems Approach to Achieving the Benefits of Artificial Intelligence in UK Defence

**Gavin Pearson[a], Phil Jolley[b], Geraint Evans[c]**
[a]Defence Science and Technology Laboratory, Porton Down, Salisbury, UK
[b]IBM Ltd, PO Box 41, North Harbour, Portsmouth, Hampshire, UK
[c]Cranfield Defence and Security Doctoral Training Centre, Defence Academy, UK

## Abstract[1]

The current resurgent interest in Artificial Intelligence (AI) has been driven by the availability of data (particularly labelled data), the democratisation of computing infrastructure and tooling, and the ability to combine these elements to create AI algorithms. Benefit is achieved once an algorithm is deployed into an operational system to achieve an operational advantage.

The ability to exploit the opportunities offered by AI within UK Defence calls for an understanding of systemic issues required to achieve an effective operational capability. This paper provides the authors' views of issues which currently block UK Defence from fully benefitting from AI technology. These are situated within a reference model for the AI Value Train, so enabling the community to address the exploitation of such data and software intensive systems in a systematic, end to end manner.

The paper sets out the conditions for success including:
• Researching future solutions to known problems and clearly defined use cases;
• Addressing achievable use cases to show benefit;
• Enhancing the availability of Defence-relevant data;
• Enhancing Defence 'know how' in AI;
• Operating Software Intensive supply chain eco-systems at required breadth and pace;
• Governance and, the integration of software and platform supply chains and operating models.

## 1 Introduction

The UK Ministry of Defence (MOD), like the US Department of Defense (DoD), has an ongoing need to adapt and flexibly employ military forces to meet changing operational needs [1, 2, 3, 4]. The US pursuit of the 'third offset' under initiatives such as Project Maven places significant emphasis on the exploitation of relevant data sets – a point reinforced by the UK's Vice Chief of the Defence Staff – using modern data analysis techniques including Artificial Intelligence (AI). Generating the notion of 'algorithmic warfare', the US intent is to generate and maintain an unassailable data exploitation position for its competitive operational advantage. AI becomes a force multiplier, with data the de facto 'Centre of Gravity'.

### 1.1 Systems Approach

Taking a 'systems approach' requires looking at an issue within the context of the system in its entirety, considering the systems behaviour as a whole, as distinct from focusing on parts of the system in isolation [5]. In accordance with principles of Boundary Critique, a critical first stage is to define the problem and the 'system under study' [6] – what is included and excluded fundamentally shapes the ability of the UK Defence enterprise to fully realise the benefits of AI. It is not just the system in which AI might be embedded, but the wider system (or even system of systems) that is required to deploy and exploit the AI technology.

The recent resurgence in AI has been driven by a near exponential growth in the availability of data (particularly labelled data), computing power, open/affordable tooling, and the ability to put these elements together to create, deploy and employ superior algorithms to deliver benefit to an enterprise. Therefore these are all parts of the system (see Figure 1), yet in pulling the elements together, there is also the realisation that Defence is itself a system of systems (which by its nature is complex and adaptive). Therefore this process will have inherent frictions which must be overcome with the appropriate degree of planning rigour.

Further there is the fact which we have all become very well aware of over the last 30 years and that is the rapid rate of technology development within Information Systems. Additionally, in the military, as in so many domains, there is uncertainty over the future situations within which military forces will be required to operate and thus the performance required from their socio-technical Information Systems (including AI). The complexities of the Contemporary Operating Environment, and the potential vagaries of that in the future, present planning conundrums – is a capability deigned to be versatile for use across much of the Spectrum

---

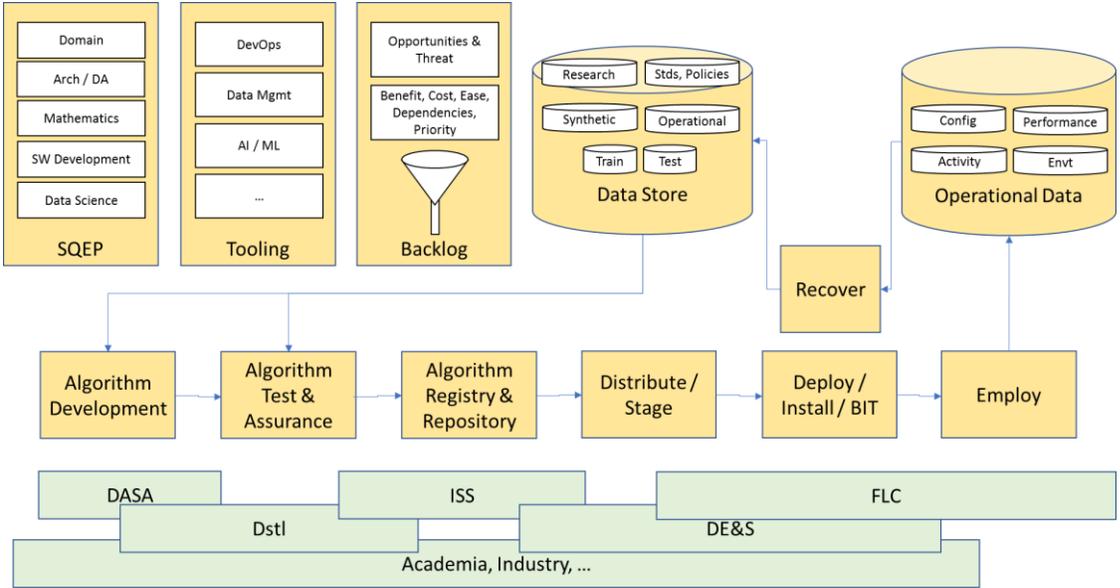

*Figure 1: Strawman AI Value Train*

of Conflict? Or is it narrowly defined for use in a specific type of operation and adversary? This uncertainty coupled with the 'democratisation of technology' means that Defence needs to act at sufficient pace to maintain Information Advantage as and when it is required along temporal, geographic and mission-specific lines. It would be unrealistic to expect it to be sustained indefinitely. UK Defence must be able to quickly generate it when needed.

Additionally the system needs to be affordable, a by no means insignificant factor in a time of strong fiscal headwinds for UK Defence. Therefore, we need the ability to evolve our AI system in an effective, timely and cost-effective manner. These four elements are illustrated in Figure 2.

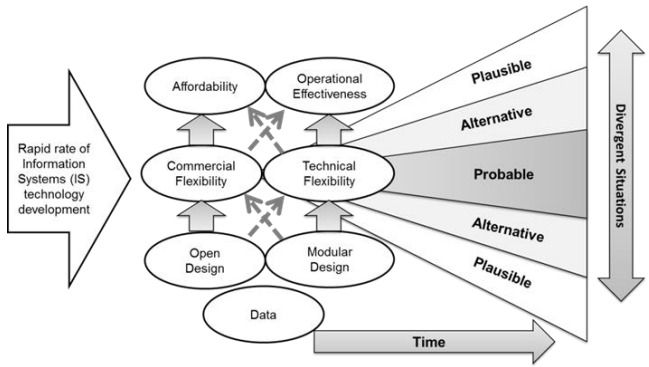

*Figure 2: Four elements of the broad context of AI*

## 2 AI immaturity: Defence Challenge to AI Technical Capabilities

One challenge with the adoption of AI within Defence is that many Defence tasks require AI capabilities which are currently immature. Examples include [7, 8]:

- Military decision-making within combat operations can be characterised as having a "high regret" if the "wrong" decision is made; so requiring a high degree of trust in any decision-making conducted by AI and the associated need for interpretability. The complexity of this challenge is compounded by the need to consider military-decision-making in different combat scenarios, operating environments and levels of command. *This is an open research question*.
- Military operations may be undertaken against an active opponent, who is deliberately attempting to deceive the AI systems; and the adversary may be using their own AI systems to control the deception operation. *This is another open research question*.
- Military operations will almost certainly be Multi-Domain in nature, with force elements active across Land, Maritime, Air, Space and Cyber domains in parallel, creating the need for AI capable of linking across highly-dispersed warfighters and agents (robotic and software) operating in complex environments *(another open research question)*.
- Military operations are frequently conducted in new settings, so there is a need for AI which can operate effectively with sparse data sets and very little training data *(another open research question)*.
- Military operations are increasingly Coalition operations in which many parties come together to achieve common aims, but operate under different sets of policies and with differing levels of trust. Thus there is a need for AI systems which can learn and reason effectively when distributed across a set of partners, whilst also retaining the ability for the UK to operate alone for sovereign purposes (*another open research question*).
- Tactical military operations are also characterised by their operation on edge computing infrastructures with limited resources, and an electro-magnetic spectrum that is both congested and contested. This is a very different

setting for the typical server farm or 'cloud' environment and again gives rise to *open research questions.*

Defence funded AI research is active in these areas, yet if it is to avoid the sins of its past, there is the need to manage stakeholder expectations very carefully, so that their demand signal for AI is pragmatic and achievable. The MOD Chief Scientific Adviser has been particularly direct in his description of AI. His description of it being 'no more than architecture, data and algorithms' was deliberate, noting that there is a danger of AI being 'over-sold' and thus unintentionally entering a 'trough of despondency' within Defence.

Finally it is worth touching on the need to match the capabilities and behaviour of AI ensembles to the type of understanding and decision-making activity which is being undertaken. Here it is worth reflecting on the very different responses needed by systems (including human organisations) when addressing open versus closed questions [9]; or when being data or hypothesis led in their attempt to understand the world [10]. Ongoing Army-sponsored research suggests that the Understanding function as doctrinally codified is itself unachievable – staffs are almost certain to never be able to fully appreciate or comprehend a situation (especially if it is in the chaotic region of the Cynefin Framework or is a 'wicked problem'), but instead must be able to interpret certain events and factors at a given point in time, or seek to predict any future change which may occur. The research suggests that understanding is more closely linked to risk mitigation and tolerance for any decision-making which follows. This is where AI could play a particularly important role in helping staffs to focus on the response they seek or, even if they do not know it initially, they need it.

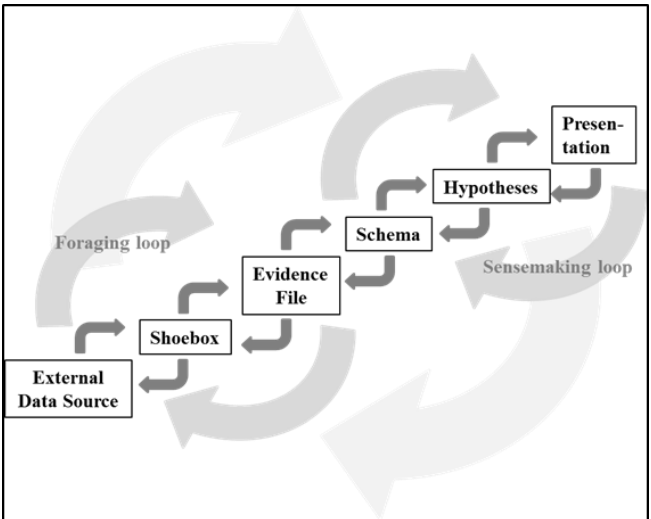

*Figure 3: Analyst Sense-Making Process [10]*

## 3 Nearer term exploitation of AI

There are, however, many ways in which AI can be used within the Defence business, as it is being used within many other businesses of a similar size and organisational complexity (e.g. within logistics and utilities maintenance). Therefore, it is worth focusing on the issues associated with nearer term exploitation of AI within Defence. This must, going forward, be based on clear and unambiguous use cases with varying degrees of complexity and ambition. The temptation to keep everything simple and in a neatly framed problem is always here. At some point, Defence stakeholders need to be bold and embrace more complex problems to investigate resolving with AI techniques.

Use of AI relies on availability of relevant data, computing infrastructure, tooling and the 'know how' to put the elements together to develop algorithms and deploy them into the hands of users. We suggest that tooling and computing infrastructures are not the elements limiting the ability to deploy AI in many areas (though the diversity of MOD computing infrastructures is a factor as discussed in section 7). Instead the critical issues are availability of and access to data, particularly labelled data (as discussed in section 4), 'know how' (as discussed in section 5) and operating a software intensive supply chain (as discussed in section 6).

Further the successful integration of AI means changing what people do and how the business operates; so it is part of a larger business change process. The paradox here is that UK Defence is a largely conservative organisation with a strong ethos of 'making do' and succeeding with limited resources. But its professionalism and strong moral component means it recognises the need to adapt and is fomenting a drive for far-reaching modernisation. As such it requires business use cases where the benefits of applying AI can be clearly articulated, and requires the development and testing of the changed business processes as well as the technology. Therefore it benefits from a senior champion who must not shy away from the difficult decisions [4].

## 4 Data

"Data is the raw material of the 21st century" [11], and is the foundational component which must be available in order to enable exploitation of AI (as without data then there is no data for an algorithm to process). Further the recent advance in AI has been driven by the availability of human labelled data which is used when training the Machine Learning component of an AI system. Therefore, it is vital that early exploitation of AI is driven by the availability of sufficient relevant data of sufficient quality, and supported by an suitable approach to creating labelled data where that is needed. Note here the point on Data and the centre of gravity – UK Defence will not fully embrace the benefits unless its data access is addressed, allowing the data to be appropriately accessed for exploitation by AI technologies. The scale of that challenge must not be underestimated.

### 4.1 Defence Data about Defence Systems

Many Defence Systems are legacy systems or are operated in a manner which means their support and maintenance is contracted out. Hence it is by no means always the case that the commercial agreement under which they were procured makes provision for their instrumentation and for the data, from any instrumentation, to be available to the wider Defence enterprise. In essence, legacy contractual decisions place what is effectively a commercial blocker to AI integration and exploitation in the Defence Equipment Program's near-term activity. Moves towards more open standards downstream provide greater opportunity and flex, yet how are operational demand signals for AI able to be met in the interim?

Further instrumenting Defence assets, so creating an effective Internet of Military Things, has implications where (i) emission control is needed in order to deny an opponent information and (ii) an opponent is actively engaging in cyber operations. This is itself acting as a key driver for AI – the Future Force hypothesis suggests much smaller physical and electronic footprints in the battlespace, which is where AI technology which may facilitate this has been identified as having particular utility in Land Command and Control (C2) studies. In addition there are other implications associated with legality, privacy and ethics.

However, adopting a Defence equivalent of the UK Government Open Data Rating scale and applying it across Defence projects would help create the conditions for success by providing Defence with access to data from its own systems. At the same time it must be asked, does Defence know where all its data is and can it be made readily available for AI testing and experimentation?

### 4.2 Data about wider world

As noted earlier, many Defence operations require data about the exterior world. Thus many Defence systems include sensors which collect data about the world. This is not simply data on the physical domain, but also includes information about the cognitive and information domains. As such data collection and analysis (inc. AI) needs to be conducted in a manner which confirms to local and UK ethical and legal constraints – including preservation of privacy and proportionality. Yet consider the potential benefit – could AI be used, for example, to unpick the digital 'fog of war' and unambiguously identify Indicators and Warnings from across data about the wider world which could, if appropriately exploited, generate Courses or Action which themselves could be accelerated and risks reduced by using other AI technologies which feed off the conditions and actions or other elements in the system (or even system of systems?).

## 5 "Know How" within Defence

There is a recognised shortage of Suitably Qualified and Experienced Personnel (SQEP) within the UK to maximise the exploitation of AI. This shortage is naturally reflected within Defence across civilian and military staff. The career management of the small pool of Defence SQEP aside, the MOD is constrained in its ability to pay more for scarce skills. However, Defence is able to compete for SQEP on other grounds.

In response to the opportunity offered by AI and need to manage 'know how', the Defence Science and Technology Laboratory (Dstl) has created an 'AI lab' which aims to be "A single pan-Dstl flagship for AI, Machine Learning and Data Science that works with suppliers and partners to establish a world-class capability in the application of AI related technologies to Defence and Security challenges" [12]. 'AI Lab' will work with similar centres of excellence, such as Project Nelson [13], being set up within the Defence Commands.

## 6 Software Intensive Supply Chain Eco-System

Experience dealing with successful Open Systems shows that it is equally important to focus on the commercial, supply chain and operating model, as it is the technical architecture [14, 15]. Additionally, any 'AI system' is clearly a software intensive system and experience teaches that the production of software is best considered a design activity rather than a production activity. Thus there is a need to adopt an operating model which employs Lean, Agile and DevOps principles and practices, noting that some capabilities will be Joint Forces Command-sponsored, whilst others will be championed and funded by the Single Services. This naturally has implications for the supply chain and commercial strategy adopted to enable the acquisition of AI within government systems. Critically it does not lend itself to the standard CADMID model of Defence acquisition.

If one accepts that a significant portion of AI software innovation occurs within Small & Medium Enterprises (SMEs) then it is necessary to design a supply chain and commercial construct that works for such SMEs. Traditionally, the large number and size of Defence standard terms and conditions which apply to procurement make it hard for an SME to engage, as understanding and complying with such a large set of terms and conditions is costly. Finding an approach which enables the effective integration of a wide and rapidly fluctuating pool of SMEs into the supply chain of such software intensive systems is a key issue which needs further investigation. It is a long-standing question in UK Defence which must be tackled head on. As a case in point, the Defence and Security Accelerator initiative has led to the emergence of some startling and potentially ground-breaking proposals and proof of concept demonstrations. Yet the pull-through to the Equipment Program is so convoluted thereafter, both from a Defence stakeholder and SME perspective, that some ideas can 'waste away.'

Furthermore, it must address how to deal with the sudden unavailability of an SME which has delivered an important micro-service. There are both technical and Intellectual Property Right (IPR) challenges. Firstly, there is the technical challenge of maintaining and, when necessary, adapting or replacing code which has been generated by an SME. Secondly, the IPR associated with both algorithms and data can, at such times, be hugely contentious. Should algorithms trained on MOD data be intellectually owned by MOD? There is the need, therefore, to differentiate between COTS algorithms used in AI tooling (by MOD and its suppliers) and those AI algorithms, and their data architectures, generated by the MOD and its suppliers.

The move to an open system, services, agile, incremental and DevOps model would mean that the average size and complexity of individual projects will fall. This will impact both major companies and SMEs, as the cost of the overheads associated with engaging with Defence will become a more significant element of each project – unless we can develop new models of engagement which are (in the jargon) 'frictionless'. Otherwise the attempt to broaden the supply pool may have entirely unintended consequences.

There also exists the need to understand the regulatory framework around development and testing of such services – particularly given the need to "harness AI for defence & security in a manner that is **moral & ethical**, reinforces **international norms** and counters **irresponsible use of AI**." This should not stymie AI exploration and the understanding of AI threats and opportunities – after all, potential UK adversaries may have a radically different moral compass – but the strictures it will force have to be sensibly acknowledged upfront.

## 7 A Fragmented Defence Information Infrastructure

Most organisations contemplating the exploitation of AI have a relatively simple information infrastructure. Unfortunately this is not the case within Defence. Defence has traditionally been in the business of either 'equipping the man' or 'manning equipment', which leads to a situation in which multiple platforms (e.g. ships, boats, planes, helicopters & vehicles) are purchased from a wide range of suppliers as complete systems. From a capability perspective, the Information Defence Lines of Development (DLOD) and essential data can exist in silos of 'splendid isolation.' Further, over time the information systems component of each such platform has become a larger and more significant part of it. Thus we have a multitude of information infrastructures provided by any number of suppliers; each of which tends to have its own supply chain, commercial model, operating model and technical architecture – this makes re-use of (micro-)services and data across these multiple platforms a challenge. The UK-based supply chain is bad enough; reconciling this for deployed forces in a theatre of operations unnecessarily adds to the Logistics and support burden.

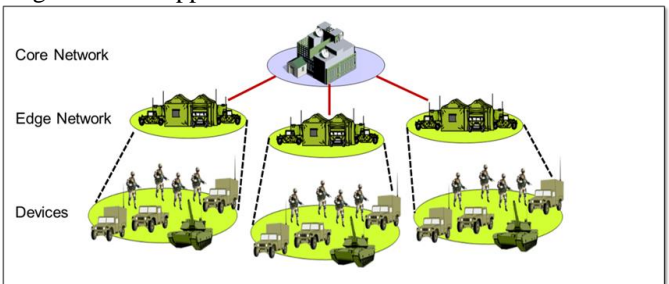

*Figure 4: Tactical Networks*

This is also why Defence has been strongly engaged within the development and testing of Open Systems approaches. However, in simple terms it means we are operating across the full set of computing paradigms (see figure 5). Importantly, as noted above, the contested nature of many Defence operations means that the communications infrastructures, and communication qualities of service, assumed by COTS solutions are not appropriate.

Worse still once we deploy forces on operations we are aiming to work with a very wide range of partners (see figure 6) – thus interoperability is now a non-discretionary requirement in UK Defence planning.

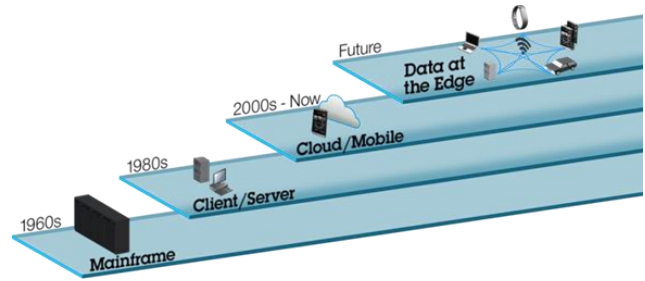

*Figure 5: Computing Paradigms (image courtesy of IBM)*

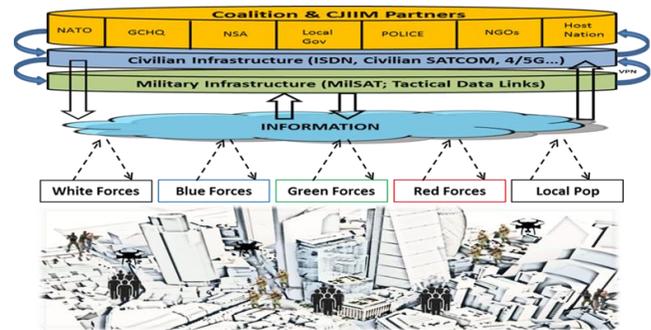

*Figure 6: Operating in the Urban Environment*

## 8 Conclusions and Recommendations

The ability to exploit the opportunities offered by AI within UK Defence requires an understanding of systemic issues required to achieve an effective operational capability.

We have set out a strawman reference model for the AI Value Train which can potentially provide a common system reference model, so enabling the community (though provision of a common lens) to address the exploitation of such data and software intensive systems in a systematic, end to end manner.

We have discussed the conditions for success including:
• Researching future solutions to known problems and clearly defined use cases;
• Addressing achievable use cases to show benefit (ones where the algorithms and computing infrastructure are mature, and data is available);
• Enhancing the availability of Defence-relevant data (outside project boundaries);
• Enhancing Defence 'know how' in AI through centres of excellence/competence and growth of expert personnel;

- Operating Software Intensive supply chain eco-systems (to access exterior 'know how' and operate at required pace));
- Governance and, the integration of software and platform supply chains and operating models.

In conclusion, we recommend a strategic approach with three main lines of action:

a) Exploit near term opportunities with an eye on the future breadth and pace required (see section 3);

b) Focus on fundamental enablers of Data, Defence 'know how' and supply chain relationships (see sections 4, 5 & 6):
   i. Develop the supply chain relationship between emerging Defence Centres of AI Excellence and wider supply chain (see section 5 & 6);
   ii. Develop a strategy to break data out from its current silos, addressing both the near-term need and longer-term opportunity (c.f. government open data rating scale) (see section 4).

c) Plan for action at required breadth, depth and pace;
   i. Maintain research on maturing AI to cope with future defence challenges and clearly defined future use cases (see section 1);
   ii. Develop and test the approach to such software intensive supply chains (see section 6);
   iii. Develop and test an approach to software intensive supply chains and their intersection with the fragmented Defence Information Infrastructure (see section 7).

## Acknowledgments and Copyright

This research was sponsored by the U.S. Army Research Laboratory and the U.K. Ministry of Defence under Agreement Number W911NF-16-3-0001. The views and conclusions contained in this document are those of the authors and should not be interpreted as representing the official policies, either expressed or implied,of the U.S. Army Research Laboratory, the U.S. Government, the U.K. Ministry of Defence or the U.K. Government. The U.S. and U.K. Governments are authorized to reproduce and distribute reprints for Government purposes notwithstanding any copyright notation hereon.